\documentclass[11pt,a4paper]{article}
\usepackage[T1]{fontenc}
\usepackage{lmodern}
\usepackage[margin=22mm]{geometry}
\usepackage{titling}
\usepackage{titlesec}
\titlespacing*{\section}{0pt}{12pt plus 2pt minus 2pt}{6pt}
\titlespacing*{\subsection}{0pt}{9pt plus 2pt minus 2pt}{4pt}
\titlespacing*{\paragraph}{0pt}{6pt plus 1pt minus 1pt}{0.8em}
\usepackage{amsmath,amssymb,booktabs,array}
\usepackage[hidelinks]{hyperref}
\usepackage{microtype}
\usepackage{enumitem}
\setlist{nosep}
\hypersetup{pdftitle={DrawOffer: Evidence-Grounded Longitudinal User-State Modeling for Career Support},
  pdfauthor={Lele Xu, Yuning Bian, Yifan Yuan}}
\title{DrawOffer: Evidence-Grounded Longitudinal User-State Modeling for Career Support}
\author{Lele Xu\thanks{Corresponding author: \texttt{jixu0090@gmail.com}.}\thanks{These authors contributed equally to this work and share first authorship.}\quad
Yuning Bian\protect\footnotemark[2]\quad Yifan Yuan\protect\footnotemark[2]\\[3pt]
Anhui University of Technology\\[2pt]
\small\texttt{jixu0090@gmail.com}, \texttt{bianyuning@gmail.com}\\
\small\texttt{yyifan355@gmail.com}}
\date{November 2, 2026}
\begin{document}
\maketitle
\begin{abstract}
Career-support assistants must distinguish user claims from evidence of current ability and withhold judgments when evidence is insufficient. DrawOffer separates language-model observation extraction from deterministic inference over source-linked records to maintain uncertain ability estimates, experience assessments, and revisable issues. We evaluate 102 fictional personas with 580 scored checkpoints, a historical full-context and Mem0 comparison, and three auxiliary knowledge-tracing datasets. On 37 test personas, DrawOffer passes 169/210 checkpoints; checkpoint-defined issue precision and recall are 0.875 and 0.636. Across three development simulations, 87.0\% of checkpoint verdicts agree, including persistent failures. A cache-only development ablation replacing answer-evidence requirements with confidence-only settlement reduces majority-vote passes from 306 to 293 of 370 and issue precision from 0.927 to 0.725. Equalizing positive source weights improves predictive metrics without changing total passes; removing decay gives mixed results. These findings support an exploratory systems contribution centered on evidence sufficiency. Developer-authored scenarios, unmatched baseline models, historical test exposure, and incomplete public artifacts limit generalization and independent reproduction.
\end{abstract}
\noindent\textbf{Keywords:} user modeling; agent memory; evidence provenance; knowledge tracing; career support; uncertainty.

\section{Introduction}
A career-support assistant accumulates heterogeneous records: resume statements, job requirements, mock-interview answers, self-reported interview reviews, and preparation activities. These records have different epistemic roles. A resume statement is a claim; an answer is an observation of performance under particular conditions; a job description specifies a target rather than evidence of the applicant's ability. Treating all three as interchangeable retrievable facts can turn an unsupported claim into an ability estimate or turn absence of evidence into a negative judgment.

We investigate how an assistant can maintain revisable user-state estimates from these records while withholding conclusions when evidence is insufficient. DrawOffer separates extraction of structured observations from deterministic state inference. Its outputs include ability bands, experience assessments, discrepancies between claims and evidence, knowledge gaps relative to goals, and the appearance or resolution of issues. These are derived judgments about a user's history, rather than verbatim recollections.

Our contributions are (i) a source-linked career-state system with explicit conditions for withholding judgments, (ii) a controlled evaluation reporting both intended behavior and failures, and (iii) paired development ablations separating evidence sufficiency from source weights and decay. The historical memory comparison and educational prediction experiments provide context, with distinct validity limits. The contribution is an applied system study; the probabilistic components use established methods.

We ask whether the system produces intended career states, how historical configurations compare, what educational sequences reveal about prediction and decay, and how simulation variation and mechanism removal affect results.

\section{Related Work}
\paragraph{Long-term assistant memory.}
LoCoMo evaluates long conversational histories using persona- and event-based conversations with human verification \cite{locomo}. LongMemEval evaluates extraction, multi-session reasoning, temporal reasoning, knowledge updates, and abstention \cite{longmemeval}. These tasks already encompass several challenges that motivate our system. Our scenario evaluation is narrower in domain and lacks their independent human verification; its purpose is to inspect state judgments in a career-support workflow.

Mem0 extracts, consolidates, and retrieves conversational memories \cite{mem0}. Zep uses temporal knowledge graphs \cite{zep}, while A-MEM organizes and revises linked memory notes \cite{amem}. Belief Memory maintains probabilistic interpretations under partial observability \cite{beliefmem}, and Hindsight distinguishes facts and beliefs within structured agent memory \cite{hindsight}. These works preclude a broad novelty claim based simply on making memory structured, temporal, revisable, or confidence-aware. DrawOffer instead emphasizes a particular latent-state task: determining what observed answers justify about ability and when evidence is sufficient to issue a user-facing judgment. The present experiments do not establish superiority over these systems. Only the historical Mem0 configuration was run as a direct memory baseline.

\paragraph{Learner modeling.}
Bayesian knowledge tracing updates a learner's mastery state from responses \cite{bkt}. Performance Factors Analysis models performance from skill-specific success and failure history \cite{pfa}, and Deep Knowledge Tracing uses recurrent sequence models \cite{dkt}. DrawOffer's Gaussian state and weighted observations apply established modeling ideas to heterogeneous records. We use educational datasets to probe predictive behavior, not to validate the semantic extraction or practical utility of the career-support system. The study does not provide an exhaustive comparison with open learner models or cognitive-diagnostic methods.

\section{Task and System}
\subsection{Sources, observations, and outputs}
Let $D_{\leq t}$ denote the sources available by time $t$. Each source has an identifier, occurrence time, and type. Observers map records to observations containing a target entity, measurement type, value, and provenance. Provenance may be an exact text span or a link to structured fields. Deterministic assembly replays valid observations in time order to construct state $S_t$.

The system distinguishes (i) what the user stated, (ii) what an observed action supports, and (iii) what the current model infers. External job requirements can define a target gap but cannot independently raise an ability estimate. Outputs may remain unassessed when direct evidence is insufficient. A low estimate and insufficient assessment are therefore different states.

Ability outputs concern knowledge points and general competencies. Experience outputs include contribution weight, role, depth, and whether particular follow-up aspects were explained clearly. Issues include gaps relative to a goal, overstated or understated claims, and unclear experience explanations. Issue state tracks appearance, persistence, and resolution rather than treating every extraction as a permanent fact.

\subsection{Observation extraction and validation}
Rule observers read deterministic fields. Model observers perform semantic tasks such as mapping a question to catalog concepts, extracting claims, and assessing experience answers. Structured responses undergo schema validation. Quote-bearing fields are checked against source text. A matching quote establishes that text exists; it does not establish semantic entailment of the resulting judgment.

The experience-answer observer requests three votes per question, each with at most two attempts, and aggregates successful votes; even one successful vote can determine the result. Failure counts must accompany voting results. M2 v11 restricts contribution weight to W1--W5 or unknown and role to participate, own, lead, solo, or unknown; invalid terms become unknown. Follow-up paraphrases allow 80 characters.

\subsection{Ability update}
The checked inference path represents ability by a Gaussian state $\theta\sim\mathcal N(\mu,\Sigma)$. An observation acts on a projection $z=u^\top\theta$, whose prior mean and variance are $m=u^\top\mu$ and $v=u^\top\Sigma u$. For a binary response $y$, discrimination $a$, and difficulty or depth threshold $b$,
\begin{equation}
 p(y=1\mid z)=\sigma(a(z-b)).
\end{equation}
A source reliability $r$ weights the likelihood:
\begin{equation}
 p(z\mid y)\propto\mathcal N(z;m,v)\,[p(y\mid z)]^r.
\end{equation}
Here $\sigma$ is the logistic function, $u$ maps an observation to ability coordinates, and $r\geq0$ tempers its influence rather than denoting a calibrated probability of correctness. A one-dimensional Laplace approximation gives mode $\hat z$ and variance $\hat v$. For $c=\Sigma u$, the rank-one update is
\begin{equation}
 \mu^+=\mu+\frac{c}{v}(\hat z-m),\qquad
 \Sigma^+=\Sigma+\frac{cc^\top}{v^2}(\hat v-v).
\end{equation}
The log-posterior derivatives are
\begin{align}
 \ell'(z)&=-(z-m)/v+ra(y-p),\\
 \ell''(z)&=-1/v-ra^2p(1-p),\qquad p=\sigma(a(z-b)).
\end{align}
Default source weights include 1.0 for real-interview observations, 0.8 for mock interviews, and 0.3 for self-reports. External material and user actions alone have zero ability-evidence weight. These are expert-set assumptions, not empirically optimal values.

Time evolution returns the estimate toward baseline mean $\mu_0$ with reference covariance $\Sigma_0$. For elapsed time $\Delta$ and stability $s_i$, set $f_i=\exp(-\Delta/s_i)$, $g_i=\sqrt{1-f_i^2}$, $F=\operatorname{diag}(f)$, and $G=\operatorname{diag}(g)$. Then
\begin{align}
 \mu'&=\mu_0+F(\mu-\mu_0),\\
 \Sigma'&=F\Sigma F+G\Sigma_0G.
\end{align}
Direct responses update stability. The mechanism weakens old evidence about current state; it is not a validated individual forgetting law.

The cache replay uses inference engine~1 and parameter set 2026.10.0. Historical business labels ``v3'' and ``v2'' identify career-system versions, not inference-engine numbers. Later factor-based, session-effect, and ordinal-error extensions are outside the scenario claims.

\subsection{Evidence sufficiency and revisable issues}
A confident posterior alone is not always enough to issue a judgment. The \texttt{settled} decision checks posterior confidence against $p_{\mathrm{decide}}=0.7$ and also considers direct-answer evidence. Depending on the target, this includes answer counts, distinct covered points, freshness, and minimum temporal spread. Other outputs retain their own assessment and coverage conditions. These gates aim to prevent a statement or a narrow observation from being generalized into a broad deficiency.

Experience explanation facets use their own weighted clear/partial/unclear evidence, temporal decay, and appearance/clearance thresholds. Thus there is no single threshold whose removal disables every evidence rule. Our ablation specifies precisely which decision function is changed.

When a source is removed or invalidated, state can be rebuilt from the remaining sources so that the withdrawn source no longer supports a conclusion. This concerns inference dependencies; it does not assert deletion from model-provider logs, caches, or all storage copies.

\section{Evaluation Design}
\subsection{Controlled career scenarios}
The expanded collection contains 102 fictional personas covering five directions---backend, frontend, AI, product, and operations---and four career stages---intern, campus entrant, mid-career, and senior. There are 581 checkpoint definitions. One requires an unavailable material observer and is excluded, leaving 580 scored checkpoints: 370 for 65 development personas and 210 for 37 test personas.

Checkpoints prescribe observable state at particular dates, including expected ability bands, required or prohibited issues, experience assessments, merging of repeated experience mentions, and state transitions. Ability and experience-band checks generally allow a one-band discrepancy; an explicitly expected insufficient-assessment state must match exactly. Issue-absence checks treat resolved and dormant issues as inactive. Supplemental-fact checks use normalized text containment or matching numeric tokens, a heuristic weaker than entailment. The development team authored the scenarios and expectations; these are not independent human gold labels.

The first round retains records for the original 25 personas and adds 77 personas simulated with the locally configured Kiro CLI model identifier \texttt{claude-haiku-5.5}. Interviewer and grading calls use product prompts with the configured observer target \texttt{deepseek-v4.1-flash}. Two additional simulations cover all 65 development personas. These are local invocation identifiers, not independently verified immutable provider snapshots. Cached identical observer requests can be reused across runs; the three runs are not independent resampling of every model call.

The 37-persona test split includes the nine test personas from the historical comparison. It is therefore not an entirely new, never-exposed test set. The expanded test split was opened on November~1, 2026. Subsequent mechanism analyses use development data only and do not tune the engine on the expanded test outcomes.

\subsection{Historical memory comparison}
The historical comparison uses 25 personas and 129 checkpoints, with 16 development personas and nine test personas containing 45 checkpoints. DrawOffer, a full-context baseline, and a Mem0 configuration receive the same underlying records. DrawOffer's observer target is \texttt{deepseek-v4.1-flash}; baseline answering uses the local CLI identifier \texttt{gpt-6.1-terra}. Model identities and budgets are unmatched, so this is a comparison of system configurations rather than a controlled memory-algorithm comparison.

The Mem0 adapter declares \texttt{mem0ai==2.2.1}, uses local BAAI/bge-small-zh-v1.5 embeddings and Qdrant storage, adds each episode as a user message in chronological order, and retrieves up to 20 memories per query. It calls \texttt{Memory.add} without explicitly disabling inference and records returned memory IDs for source deletion. This is not evidence that Mem0 only appends memories or cannot update them. Internal historical update actions and source ownership after cross-source updates have not been fully reconstructed.

\subsection{Public knowledge-tracing experiments}
Auxiliary experiments use ASSISTments 2009--2010 \cite{assist}, EdNet-KT1 \cite{ednet}, and Junyi Academy \cite{junyi}. Retained sizes are 4,027 students/282,969 interactions, 5,000/703,306, and 5,000/787,452, respectively. Student-level five-fold splits use seed~0. Predictions precede assimilation of the current outcome.

ASSISTments drops rows with missing skills, groups multiple skill rows for the same student and order identifier into one interaction, and retains students with at least two interactions. Since the loader does not provide real timestamps, interactions receive equal one-minute spacing and this dataset is not used as evidence about forgetting. EdNet derives correctness from the question answer key and skills from tags; missing/untagged questions are skipped. Students require at least 20 retained interactions, with the first 500 retained. Although the helper invokes a seeded permutation, it sorts the full candidate list before selection: the implemented EdNet subset is the first 5,000 eligible files in filename order, not a uniform random sample. Junyi samples 5,000 eligible users with seed~0, requires at least 20 interactions, retains the first 500, and uses the exercise as both problem and skill. The loader reads the provided correctness flag and timestamps.

BKT uses pyBKT EM with five fits per skill and averages skill probabilities for multi-skill items. The PFA variant uses $\log(1+n)$ success/failure counts and logistic regression ($C=1$, at most 2,000 iterations), rather than the original raw-count specification. IRT fits L2-penalized Rasch difficulties on training students and updates test-student ability online. DKT uses a 100-unit LSTM, Adam ($10^{-3}$), batch size 32, gradient clipping at 1, and at most 30 epochs, with patience 3 on a 10\% student-level validation subset of the training fold. It predicts multi-skill items by applying a sigmoid to the mean skill logit. These are local model-family configurations, not published leaderboard reproductions.

The default engine fits no training difficulty; the skill-difficulty variant matches a smoothed training correctness rate per skill. The problem-difficulty control uses the same training Rasch fit as the IRT baseline. The no-forgetting variant removes state drift from the skill-difficulty configuration. EdNet and Junyi after-gap subsets require all skills of the current item to have been observed before and at least one day to have elapsed since each skill's last occurrence. Reported AUCs are means of fold AUCs, including after-gap AUCs.

\subsection{Metrics and statistical scope}
Overall scenario pass rate is the fraction of scored checkpoints passed. Issue metrics are defined over explicit positive and negative checkpoints: $\mathrm{TP}$ counts passing issue-present checks, $\mathrm{FP}$ counts failing issue-absent checks, precision is $\mathrm{TP}/(\mathrm{TP}+\mathrm{FP})$, and recall is $\mathrm{TP}$ divided by the number of positive checks. These are checkpoint-defined metrics, not precision and recall over every possible issue the system could emit. The evidence-gate metric is the pass rate of issue-absent checks. Issue-state checks separately test transitions.

Prediction uses AUC and log loss. Historical AUC-difference intervals resample question units and do not account fully for dependence between questions from the same persona. We do not interpret them as persona-level significance tests. Development ablation pass rates first take a majority verdict across the three runs for each checkpoint. Paired intervals then resample the 65 personas with replacement 10,000 times, keeping all of a persona's majority verdicts together (seed 20261101); they do not resample runs or model calls. Predictive AUC and log loss instead pool response predictions across the three runs. Intervals are exploratory pointwise percentile intervals without multiplicity correction.

\section{Results}
\subsection{Expanded career scenarios}
The test split passes 169/210 checkpoints (80.5\%); the first development round passes 302/370 (81.6\%). Table~\ref{tab:career} compares the business-system versions on the test split. Better issue recall coexists with lower ability-band and experience-assessment performance. The 223 test response predictions give AUC .7155 and log loss .5917, but these do not establish probability calibration. The overall pass rate weights checkpoints equally, not personas or assertion types.
\begin{table}[htbp]\centering
\caption{Expanded scenario test results (37 personas, 210 checkpoints). Check rows report pass rates; precision/recall use the checkpoint definitions in Section~4.4. The predecessor is a business-system baseline.}\label{tab:career}
\begin{tabular}{lrr}\toprule
Metric & DrawOffer v3 & Predecessor v2\\\midrule
Ability-band checks & .913 & .957\\
Checkpoint-defined issue precision & .875 & .818\\
Checkpoint-defined issue recall & .636 & .273\\
Experience-assessment checks & .581 & .613\\
Inactive-issue checks & .962 & .975\\
Issue-state transitions & .125 & .000\\
Experience merging & 1.000 & 1.000\\
Readiness checks & 1.000 & 1.000\\
Supplemental-fact checks & .000 & .000\\
Next-response AUC & .7155 & .6883\\
Next-response log loss & .5917 & .6233\\\bottomrule
\end{tabular}\end{table}

By direction, test passes are 31/37 (backend), 32/40 (frontend), 43/50 (AI), 29/41 (product), and 34/42 (operations). By stage, they are 21/24 (intern), 47/49 (campus), 51/66 (mid-career), and 50/71 (senior). Different checkpoint mixes and small denominators prevent causal interpretations of these differences.

Experience-explanation failures are substantial: only 1/6 checks requiring an unclear-explanation issue pass, and 0/5 checks of subsequent resolution pass. Inspection identifies both implementation and benchmark problems. Real-interview reviews merge follow-up text into one answer, leaving the facet observer without the expected follow-up structure; some simulated conversations never elicit the intended facet. These outcomes should not be attributed solely to the inference algorithm.

\subsection{Historical general-memory comparison}
Table~\ref{tab:memory} shows that DrawOffer does not lead overall checkpoint accuracy. The historical test AUC point estimate is higher, but its difference from full-context answering has interval $[-.0697,.1460]$ and its difference from Mem0 has interval $[-.0435,.1745]$. Both cross zero and use the historical question-level resampling procedure.
\begin{table}[htbp]\centering
\caption{Historical comparison on 25 personas, including nine test personas. Different model configurations limit attribution.}\label{tab:memory}
\begin{tabular}{lrrr}\toprule
Configuration & All passes & Test passes & Test AUC\\\midrule
DrawOffer & 105/129 & 40/45 & .8093\\
Full context & 109/129 & 40/45 & .7677\\
Mem0 adapter & 107/129 & 38/45 & .7433\\\bottomrule
\end{tabular}\end{table}

Exact-source/quote checks pass for 520/540 original engine conclusions. Including derived states yields 673/770, of which 153/230 concern derived items. These denominators are different and must not be combined into a single traceability claim. Rewritten retrieved memories may not preserve verbatim source spans, so exact-string checking disadvantages such configurations. These numbers measure source/quote presence, not truthfulness or semantic support.

\subsection{Auxiliary knowledge tracing}
Table~\ref{tab:kt} reports five-fold mean AUC. Difficulty calibration changes the engine's performance substantially. No configuration dominates every baseline across datasets.
\begin{table}[htbp]\centering
\caption{Local knowledge-tracing configurations; five-fold mean AUC.}\label{tab:kt}
\begin{tabular}{lrrr}\toprule
Configuration & ASSISTments & EdNet & Junyi\\\midrule
Engine, no trained difficulty & .7010 & .6521 & .7397\\
Engine, skill difficulty & .7189 & .6765 & .7863\\
Engine, problem difficulty & .7676 & .7577 & .7903\\
BKT & .7173 & .6730 & .7868\\
PFA & .7213 & .6783 & .7903\\
IRT & .7623 & .7626 & .7768\\
DKT & .7514 & .7033 & .7960\\\bottomrule
\end{tabular}\end{table}

On EdNet after-gap interactions, the skill-difficulty engine achieves .6093 AUC versus .6031 without forgetting, .5886 for BKT, .5815 for PFA, and .6185 for DKT. On Junyi, the corresponding engine values are .7659 and .7629, versus .7390, .7432, and .7499 for BKT, PFA, and DKT. The decay gains of .0062 and .0030 are descriptive and local to these configurations. Five folds with a consistent direction do not constitute a significance test. These results do not validate career-state judgments or a general model of human forgetting.

\subsection{Stability, withdrawal, and computation}
The three development rounds pass 302, 297, and 307 of 370 checkpoints. Majority voting passes 306/370 (82.7\%). Verdicts agree in all three rounds for 322/370 checkpoints (87.0\%): 276 always pass and 46 always fail; 48 vary. Agreement therefore includes persistent errors. This experiment combines simulation and observer variation and does not isolate either component's randomness. Rate-limit-induced missing outputs were completed under the original settings; final cache-only replay has zero misses and no new calls.

With fixed observations, the historical 25-persona determinism test reports identical states across five in-process repetitions and two fresh processes. In 85 source-withdrawal cases, none of the 176 tracked dependent conclusions survives, and no residual source observations or conclusions are reported. These are tested dependency properties rather than general correctness or comprehensive erasure guarantees.

Synthetic timing uses six questions per session, five learners per scale, and three repeats. At 3,000 answers, inference engine~1 full reconstruction takes median .2767 seconds and P95 .7985 seconds. The distinct engine~2 takes median 1.6033 seconds for full reconstruction and .0384 seconds for adding one session incrementally; incremental and full results agree for the five tested learners. The incremental number must not be attributed to engine~1. The report records catalog 2026.10.7, Python 3.13.13, NumPy 2.5.4, and Darwin arm64. These are small pure-computation measurements excluding model extraction, network, and database latency, not end-to-end service benchmarks.

\subsection{Development mechanism ablation}
Before inspecting ablation outputs, we recorded the interventions, development-only scope, and bootstrap procedure. All configurations reuse the same three sets of source records and structured-observer cache. The full configuration reproduces historical per-round failed-checkpoint sets, aggregate metrics, majority failures, and predictive results. Every replay has zero model calls, zero cache misses, and no failed model requests. The replay constructs no model gateway or database client.

Three interventions are applied in isolated Python processes without changing product defaults:
\begin{enumerate}
\item \textbf{Equal positive reliability:} replace every positive source-class weight by 1, including mock/real experience weights; retain zero-weight non-evidence categories and grader-version factors.
\item \textbf{No decay:} set ability and experience stability and explanation-facet half-life to infinity. Keep actual dates, freshness rules, and minimum temporal evidence-spread conditions.
\item \textbf{Confidence-only settlement:} replace \texttt{Posterior.settled} with a mapped-node posterior confidence test at $p_{\mathrm{decide}}=.7$. Remove its answer-count, covered-point, and temporal-spread requirements; retain other business, coverage, and grading rules.
\end{enumerate}
These are scoped interventions, not removal of every reliability, temporal, or evidence-related operation in the system.

\begin{table}[htbp]\centering\small
\caption{Development majority-vote ablation. Delta is ablation minus full pass rate, in percentage points. Intervals cluster by persona.}\label{tab:ablation}
\begin{tabular}{lrrr}\toprule
Configuration & Passes /370 & Delta & Exploratory 95\% interval\\\midrule
Full & 306 & 0 & ---\\
Equal positive reliability & 306 & 0.00 & $[-1.08,1.09]$\\
No decay & 303 & $-.81$ & $[-1.94,.28]$\\
Confidence-only settlement & 293 & $-3.51$ & $[-6.05,-1.33]$\\\bottomrule
\end{tabular}\end{table}

Confidence-only settlement improves one checkpoint and worsens 14; issue precision falls from .9268 to .7255, recall from .5938 to .5781, and issue-state pass rate from .2727 to .1818. The result supports answer-evidence requirements on these controlled tasks, not the calibration of the confidence threshold itself. However, developer-authored issue-absence checkpoints explicitly reward restraint; independent judgment is needed before claiming real utility.

Equal reliability improves two checkpoints and worsens two. Its next-response AUC increases from .7867 to .7988 and log loss decreases from .5418 to .5208. The selected reliability weights are therefore not supported as a predictive improvement by this experiment. No decay improves one checkpoint and worsens four; its AUC is .7862 and log loss .5389, giving mixed predictive directions and a pass-rate interval crossing zero. We retain the mechanisms as design choices but do not claim that both improve performance.

\section{Discussion and Limitations}
\paragraph{What is supported.}
The strongest local result is the deterioration under confidence-only settlement. It suggests evaluating the permission to issue a judgment separately from predictive accuracy. This does not establish superiority over memory systems, optimal source weights, calibrated ability probabilities, or employment benefits.

\paragraph{Evaluation dependence.}
Developer-authored expectations may favor the system's intended rules, especially restraint on issue-absence checks. Simulators may fail to elicit intended evidence. Neither semantic support nor ability truth has independent annotation. Historical test exposure, shared personas and prompts, and cache reuse limit independence; checkpoint precision also omits issues outside the predefined checks.

\paragraph{Baseline and version limits.}
Unmatched models and incomplete immutable provider records prevent attributing historical differences to memory algorithms. Exact-quote tests are not neutral toward rewriting. The 25- and 102-persona experiments cannot form one controlled leaderboard. Cache replay verifies the current reconstruction, not the entire historical environment or newer engine extensions.

\paragraph{Transfer and selection limits.}
Educational correctness sequences differ from interviews in what they measure and how errors arise. EdNet's implemented filename-ordered selection can introduce selection bias. Junyi uses exercise identity as skill, and ASSISTments supplies no real temporal evidence in this loader. The KT tests are auxiliary mechanism probes under local preprocessing, not validation of the full career-support pipeline.

\paragraph{Known system weaknesses.}
Experience-explanation and resolution checks remain weak, in part because of missing follow-up structure. Quote presence is weaker than semantic entailment. Fixed source weights and temporal settings may misrepresent individual users. The system should expose uncertainty and evidence rather than present inferred deficits as verified personal facts.

\section{Data, Artifacts, and Responsible Use}
Career personas are fictional; the study does not use real resumes or contact details. Educational datasets are existing public research resources; raw datasets are not redistributed with this manuscript. Public availability does not imply unrestricted redistribution, and the original access and license terms remain applicable.

Evaluation code and aggregate results are at \url{https://github.com/ql-link/DrawOffer-Eval}. The core engine is private; generated conversations and model caches are not distributed. Public synchronization of the expanded personas and paper-specific supplement remains unverified. The repository therefore cannot independently reproduce all experiments. Local artifacts preserve ablation aggregates, per-persona verdicts, interventions, and input hashes; hashes establish identity, not availability or semantic validity.

The system is intended for user-facing career preparation, not automated hiring or selection. A real-user study would require appropriate consent, access controls, withdrawal procedures, and applicable ethical review. No claim is made that these experiments establish fairness across demographic groups, real-world ability, or legal compliance.

\paragraph{Use of AI tools.}
AI tools assisted with experiment scripting, analysis, literature lookup, translation, and drafting. Models also generated fictional interactions and extracted observations as described above. AI-generated edits do not constitute independent peer review; the human authors are responsible for verifying and approving the submitted claims, citations, and text.

\section{Conclusion}
DrawOffer separates source records, structured observations, and uncertain user-state judgments. Controlled scenarios expose incomplete experience and issue-transition behavior. Development ablations support evidence requirements locally, while source-weight and decay results remain negative or mixed. The study frames evidence sufficiency as a distinct evaluation target for longitudinal assistants, subject to the stated reproducibility and external-validity limits.

\end{document}